\documentclass{article}

\usepackage{microtype}
\usepackage{graphicx}
\usepackage{booktabs}
\usepackage{multirow}
\usepackage{makecell}
\usepackage{amsmath}
\usepackage[table]{xcolor}
\usepackage{hyperref}
\usepackage{url}

\usepackage[accepted]{icml2026}

\newcommand{\operaCT}{\textsc{Opera-CT}}
\newcommand{\operaCE}{\textsc{Opera-CE}}
\newcommand{\operaGT}{\textsc{Opera-GT}}
\newcommand{\MtwoD}{\textsc{M2D+Resp}}
\newcommand{\HeAR}{\textsc{HeAR}}
\newcommand{\std}[1]{{\scriptsize$\,{\pm}$#1}}

\icmltitlerunning{Cough Regression Benchmark for Respiratory Acoustic FMs}

\begin{document}
\twocolumn[
  \icmltitle{Beyond Classification: A Cough Regression Benchmark\\
    for Respiratory Acoustic Foundation Models}
  \begin{icmlauthorlist}
    \icmlauthor{Mayur Sanap}{aff2,ncsu}
    \icmlauthor{Prasanna Desikan}{aff2}
    \icmlauthor{Edgar Lobaton}{ncsu}
  \end{icmlauthorlist}
  \icmlaffiliation{aff2}{Centific Global Solutions Inc.}
  \icmlaffiliation{ncsu}{Department of Electrical and Computer
    Engineering, North Carolina State University, Raleigh, NC, USA}
  \icmlcorrespondingauthor{Mayur Sanap}{msanap@ncsu.edu}
  \icmlkeywords{respiratory acoustics, foundation models, regression,
    cough analysis, cross-dataset generalisation}
  \vskip 0.3in
]

\printAffiliationsAndNotice{}

\begin{abstract}
Respiratory acoustic foundation models~(FMs) excel at cough
classification, yet their ability to predict continuous health
quantities from cough audio remains largely unexplored,
despite the clinical value of passive age, BMI, and
disease-probability estimation in settings where physical
measurements are unavailable.
We introduce the multi-model, multi-target cough regression
benchmark evaluating five FMs (\operaCT{}, \operaCE{}, \operaGT{},
\HeAR{}, \MtwoD{}) across six targets on three datasets under
subject-disjoint protocols, comparing linear, MLP-small, and
full MLP regression heads.
MLP-small beats the mean-predictor baseline on all tasks and
linear probing in 23 of 30 model$\,\times\,$task cases, with
full MLP overfitting on small clinical data but recovering on
larger sets, revealing a dataset-size $\times$ head-capacity
trade-off.
\HeAR{} leads within-dataset age regression on Coswara
(9.12\,yr MAE); its CIDRZ result is excluded from headline
claims owing to possible \HeAR{}--CIDRZ pretraining overlap.
\operaGT{} is favored over \operaCT{} on age in all three
datasets, with the CIDRZ margin within seed variance,
extending a generative-pretraining advantage from breath to
cough. \HeAR{} and \MtwoD{} reach near-full performance at
$N$\,=\,50 samples while OPERA models require $N$\,=\,400.
Cross-dataset transfer is strongly asymmetric as large diverse data
generalises to small clinical populations
(CoughVID$\,\to\,$CIDRZ: $-0.17$\,yr) but not vice versa
(CIDRZ$\,\to\,$Coswara: $+2.43$\,yr, $+26.6\%$).
\end{abstract}

\begin{figure*}[t]
  \centering
  \includegraphics[width=\textwidth]{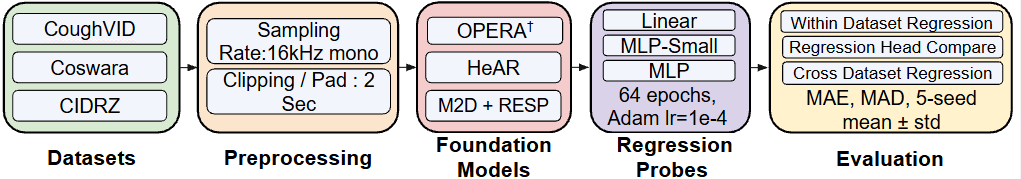}
  \caption{Cough regression benchmark pipeline. All audio is
    resampled to 16\,kHz mono and padded/trimmed to 2\,s.
    Five frozen encoders feed three regression probes and outputs
    cover three evaluation regimes.
    \textsc{Opera}$^\dagger$ comprises \operaCT{}, \operaCE{},
    and \operaGT{}.}
  \label{fig:pipeline}
\end{figure*}

\section{Introduction}

Cough acoustics encode physiological state beyond categorical disease
labels, with spectrotemporal features reflecting airway geometry,
respiratory muscle strength, and mucosal viscosity, all of which
covary quantitatively with age, body composition, and disease
severity~\citep{sharan2018predicting,xu2022forced,rudraraju2020cough}.
In low- and middle-income countries (LMICs) where respiratory disease
burden is highest~\citep{who2021tb}, birth records, weighing
equipment, and radiological capacity are often unavailable, 
making passive cough-based estimates of age and BMI actionable
proxies for triage in settings where physical measurements are
unavailable.

Foundation models (FMs) pretrained on large unlabeled audio corpora
learn task-agnostic embeddings that transfer efficiently via linear
probing~\citep{operabenchmark,baur2024hear,niizumi2025m2d},
reducing the labeled-data burden in clinical audio AI.
The three leading respiratory FM families,
OPERA~\citep{operabenchmark}, \HeAR{}~\citep{baur2024hear}, and
\MtwoD{}~\citep{niizumi2025m2d}, have been benchmarked extensively on
classification, but their regression capability is almost entirely
uncharacterised.

The \HeAR{} paper~\citep{baur2024hear} evaluates age and BMI
regression from cough on the CIDRZ and Coswara~\citep{coswara}
datasets, but restricts the analysis to a single model with a fixed
linear probe and per-device splits, with no multi-model comparison,
non-linear head evaluation, or cross-dataset generalisation.
The OPERA benchmark~\citep{operabenchmark} includes regression tasks
(spirometry estimation) exclusively from deep-breath and vowel sounds,
not from cough.
\MtwoD{}~\citep{niizumi2025m2d} has never been evaluated on cough
regression task.

We address these gaps with five contributions.
\begin{enumerate}
\item \textbf{Multi-model, multi-target cough regression benchmark}
evaluating five FMs across six targets (age, BMI, X-ray abnormality,
TB probability) on three datasets under subject-disjoint protocols,
with MAE reported alongside the mean-predictor baseline (MAD).
\item \textbf{Regression head comparison} showing MLP-small wins
23 of 30 model$\,\times\,$task combinations.
Full MLP overfits on small clinical data ($+0.53$\,yr for
\MtwoD{} on CIDRZ) but recovers on larger ones.
\item \textbf{Generative pretraining advantage} where \operaGT{}
outperforms \operaCT{} on age regression in all three datasets
(CIDRZ $10.49$ vs.\ $10.52$\,yr, Coswara $10.16$ vs.\ $10.25$\,yr,
CoughVID $9.62$ vs.\ $9.79$\,yr; the CIDRZ gap lies within seed
variance), extending the breath-regression finding
of~\citet{operabenchmark} to cough.
\item \textbf{Cross-dataset transfer asymmetry} where large
web-collected data generalises to small clinical populations
(CoughVID$\,\to\,$CIDRZ $-0.17$\,yr) while the reverse fails
(CIDRZ$\,\to\,$Coswara $+2.43$\,yr, $+26.6\%$).
\item \textbf{Low-data regime analysis} showing \HeAR{} and
\MtwoD{} reach near-full performance at $N$\,=\,50 samples while
OPERA models require $N$\,=\,400, indicating pretraining corpus
diversity determines low-data regression performance.
\end{enumerate}

The rest of this paper is organized as follows. Section~\ref{sec:design} describes the benchmark design.
Section~\ref{sec:results} presents results and
Section~\ref{sec:conclusion} summarises findings and future directions.

\section{Benchmark Design}
\label{sec:design}

We evaluate five frozen FMs on three cough datasets under
a unified pipeline (Figure~\ref{fig:pipeline}), resampling
all audio to 16\,kHz mono and padding or trimming to 2\,s,
with embeddings extracted once and shared across all regimes.

\subsection{Datasets and Tasks}
\label{subsec:datasets}

Table~\ref{tab:datasets} summarises the three datasets and six
regression targets. Subject-disjoint splits (64/16/20\%) are used
for CIDRZ and Coswara; CoughVID reuses the official UUID-level
split (train\,=\,3050, val\,=\,1019, test\,=\,2789).

\textbf{CIDRZ}~\citep{baur2024hear,cidrz} ($N$\,=\,1049) contains
smartphone-recorded volitional coughs from patients at a TB clinic
in Zambia, with labels from clinical assessments covering age
($37.1 \pm 12.9$\,yr), BMI ($21.6 \pm 5.3$\,kg/m$^2$), chest X-ray
abnormality ($0.46 \pm 0.35$), and TB probability ($0.62 \pm 0.24$).
The X-ray abnormality and TB targets are continuous derived scores,
not the binary radiological and microbiological diagnoses; we
therefore treat them as score-reproduction tasks, not
clinical-endpoint prediction~\citep{cidrz}.

\textbf{Coswara}~\citep{coswara} ($N$\,=\,2560) consists of
shallow-cough recordings collected remotely via a web application
across India during the COVID-19 pandemic, with self-reported age
labels ($35.1 \pm 13.9$\,yr, range 7--87).

\textbf{CoughVID}~\citep{coughvid} ($N$\,=\,6858) comprises
volitional cough recordings submitted globally via smartphone
with continuous age labels ($34.5 \pm 12.7$\,yr, range 5--97)
and the largest training set (3,050 samples).

\begin{table}[t]
\centering
\caption{Dataset and task summary. Prob.\,=\,prob.\ in $[0,1]$.}
\label{tab:datasets}
\setlength{\tabcolsep}{4pt}
\small
\begin{tabular}{llll}
\toprule
Dataset & Task & Unit & $N$ \\
\midrule
\multirow{4}{*}{\makecell[l]{CIDRZ\\\citep{cidrz}\\Zambia}}
  & Age            & yr       & \multirow{4}{*}{1049} \\
  & BMI            & kg/m$^2$ & \\
  & X-ray Abn.     & prob.    & \\
  & TB probability & prob.    & \\
\midrule
\makecell[l]{Coswara\\\citep{coswara}\\India}
  & Age & yr & 2560 \\
\midrule
\makecell[l]{CoughVID\\\citep{coughvid}\\Global}
  & Age & yr & 6858 \\
\bottomrule
\end{tabular}
\end{table}

\begin{table}[t]
\centering
\caption{Foundation models. All receive 2\,s at 16\,kHz.
RC\,=\,respiratory clips.
$^\dagger$\,8.18\,s positional grid; padded to 2\,s.}
\label{tab:models}
\small
\setlength{\tabcolsep}{3pt}
\begin{tabular}{llrl}
\toprule
Model & Architecture & Dim & Pretraining \\
\midrule
\operaCT{} & Contrastive Trans.       & 768  & 136K RC \\
\operaCE{} & Contrastive CNN          & 1280 & 136K RC \\
\operaGT{} & Generative MAE$^\dagger$ & 384  & 136K RC \\
\HeAR{}    & ViT-L MAE                & 512  & 313M health \\
\MtwoD{}   & Masked Mod.\ + Resp      & 3840 & AudioSet+RC \\
\bottomrule
\end{tabular}
\end{table}


\begin{table*}[t]
\centering
\caption{Within-dataset MAE (MLP-small, 5-seed mean\,$\pm$\,std).
MAD\,=\,mean absolute deviation (naive mean-predictor baseline);
best/MAD\,=\,ratio of best MAE to MAD (lower is better;
$<$0.90 indicates meaningful signal above chance).
\textbf{Bold} = best MAE per row, excluding $^{\dagger}$-flagged cells.
$^{\dagger}$\HeAR{} CIDRZ results are potentially contaminated
(CIDRZ may appear in \HeAR{} pretraining) and are excluded from
leader claims.}
\label{tab:within}
\begin{tabular}{llccccccccc}
\toprule
Task & Unit & MAD & \operaCT{} & \operaCE{} & \operaGT{} & \HeAR{} & \MtwoD{} & best/MAD \\
\midrule
CIDRZ Age    & yr    & 10.35 & 10.52\std{0.08} & 10.51\std{0.09} & 10.49\std{0.07} & 10.29$^{\dagger}$\std{0.04} & \textbf{10.40}\std{0.05} & 0.99 \\
CIDRZ BMI    & kg/m² &  3.74 & \textbf{3.60}\std{0.01}  & \textbf{3.60}\std{0.01}  & 3.67\std{0.01}  & 3.60$^{\dagger}$\std{0.02}  & 3.63\std{0.02}  & 0.96 \\
CIDRZ Abn    & prob  & 0.325 & 0.327\std{0.001}& 0.325\std{0.001}& \textbf{0.316}\std{0.001}& 0.328$^{\dagger}$\std{0.001}& 0.320\std{0.004} & 0.97 \\
CIDRZ TB     & prob  & 0.205 & \textbf{0.189}\std{0.001}& 0.191\std{0.000}& 0.190\std{0.000}& 0.188$^{\dagger}$\std{0.001}& 0.192\std{0.001} & 0.92 \\
Coswara Age  & yr    & 11.31 & 10.25\std{0.02} & 10.44\std{0.01} & 10.16\std{0.04} & \textbf{9.12}\std{0.07}  & 9.58\std{0.06}  & \textbf{0.81} \\
CoughVID Age & yr    & 10.29 &  9.79\std{0.01}  & 9.88\std{0.02}  & 9.62\std{0.03}  & \textbf{9.61}\std{0.02}  & 9.79\std{0.02}  & 0.93 \\
\bottomrule
\end{tabular}
\end{table*}

\subsection{Foundation Models}
\label{subsec:models}

We evaluate five frozen FMs spanning three pretraining paradigms:
respiratory-specific SSL (OPERA), large-scale health audio
masked autoencoding (\HeAR{}), and general audio masked modelling
with respiratory fine-tuning (\MtwoD{}).
Table~\ref{tab:models} summarises the five encoders.

\operaCT{} and \operaCE{}~\citep{operabenchmark} are contrastive
models on 136K respiratory clips, differing in architecture
(Transformer vs.\ EfficientNet CNN) and dimension (768-d vs.\
1280-d).
\operaGT{}~\citep{operabenchmark} is a generative masked
autoencoder on the same corpus with an 8.18\,s positional grid and zero-padded inputs.
\HeAR{}~\citep{baur2024hear} is a ViT-L masked autoencoder
pretrained on 313M health audio clips, the largest corpus and
the only one covering clinical audio beyond respiratory sounds.
\MtwoD{}~\citep{niizumi2025m2d} combines masked spectrogram
prediction on AudioSet with respiratory fine-tuning, with features mean-pooled to a single 3840-d vector per clip.
All encoders are kept frozen and embeddings are extracted once,
reused across all heads and evaluation regimes.

\subsection{Regression Heads and Training}
\label{subsec:heads}

We compare three heads applied to frozen embeddings.
{Linear} ($d_{\text{feat}} \to 1$) is the standard linear
probing baseline~\citep{operabenchmark,baur2024hear}.
{MLP-small} uses a 256-unit bottleneck with ReLU and 0.3
dropout ($d_{\text{feat}} \to 256 \to 1$), making it
embedding-size-agnostic and comparable across all FMs.
{MLP} is full-width ($d_{\text{feat}} \to d_{\text{feat}}
\to 1$), yielding $\sim$15M hidden parameters for \MtwoD{},
expected to overfit on small datasets and recover on larger ones.
All heads use Adam (lr\,=\,$10^{-4}$, L2\,=\,$10^{-5}$, batch 64),
LR decay 0.97/epoch, MSE loss, and early stopping (patience 10)
on validation MAE for up to 64 epochs.
Results are mean\,$\pm$\,std over 5 seeds.

\subsection{Evaluation Regimes and Metrics}
\label{subsec:eval}

We report MAE in natural units (years, kg/m$^2$, probability)
alongside the mean absolute deviation (MAD) of each label
distribution as a naive mean-predictor baseline.
MAE\,$<$\,MAD confirms learning above chance and best MAE/MAD
quantifies signal strength, since a model collapsing to the
population mean can appear effective with no patient-level
prediction. Lower MAE is better throughout.

{Within-dataset regression} evaluates all six targets on
held-out test splits (Table~\ref{tab:within}).
{Regression head comparison} covers 90
model$\,\times\,$task$\,\times\,$head combinations to characterise
the dataset-size $\times$ head-capacity interaction
(Table~\ref{tab:full_heads}).
{Cross-dataset transfer} trains MLP-small on one dataset
and evaluates on all others without adaptation across six age
transfer conditions (Table~\ref{tab:cross})

\begin{table*}[t]
\centering
\caption{Full three-head comparison (MAE, 5-seed mean).
Age/BMI in natural units; abnormality and TB in probability
units $[0,1]$. oCT/oCE/oGT\,=\,\operaCT{}/\operaCE{}/\operaGT{};
M2D\,=\,\MtwoD{}. \textbf{Bold} = best head per
model$\,\times\,$task.}
\label{tab:full_heads}
\resizebox{\textwidth}{!}{%
\begin{tabular}{lccccc|ccccc|ccccc}
\toprule
 & \multicolumn{5}{c|}{Linear}
   & \multicolumn{5}{c|}{MLP-small}
   & \multicolumn{5}{c}{MLP} \\
Task
  & oCT & oCE & oGT & HeAR & M2D
  & oCT & oCE & oGT & HeAR & M2D
  & oCT & oCE & oGT & HeAR & M2D \\
\midrule
CIDRZ Age
  & \textbf{10.47} & 10.55 & \textbf{10.44} & 10.58 & 10.63
  & 10.52 & \textbf{10.51} & 10.49 & \textbf{10.29} & \textbf{10.40}
  & 10.57 & 10.70 & 10.54 & 10.39 & 10.93 \\
CIDRZ BMI
  & 3.64 & 3.62 & \textbf{3.67} & 3.64 & 3.68
  & \textbf{3.60} & \textbf{3.60} & \textbf{3.67} & \textbf{3.60} & \textbf{3.63}
  & \textbf{3.60} & \textbf{3.60} & \textbf{3.67} & 3.61 & 3.91 \\
CIDRZ Abn
  & \textbf{0.326} & \textbf{0.324} & 0.317 & 0.331 & 0.316
  & 0.327 & 0.325 & \textbf{0.316} & \textbf{0.328} & 0.320
  & 0.331 & 0.329 & \textbf{0.316} & 0.332 & \textbf{0.315} \\
CIDRZ TB
  & \textbf{0.189} & \textbf{0.190} & 0.191 & 0.191 & 0.194
  & \textbf{0.189} & 0.191 & \textbf{0.190} & \textbf{0.188} & \textbf{0.192}
  & 0.191 & 0.193 & 0.191 & 0.189 & 0.199 \\
Coswara Age
  & 10.25 & 10.46 & 10.25 & 9.50 & 9.98
  & 10.25 & \textbf{10.44} & 10.16 & \textbf{9.12} & \textbf{9.58}
  & \textbf{10.24} & 10.49 & \textbf{10.14} & 9.26 & 9.90 \\
CoughVID Age
  & 9.81 & \textbf{9.88} & 9.71 & 9.78 & 9.86
  & \textbf{9.79} & \textbf{9.88} & 9.62 & \textbf{9.61} & \textbf{9.79}
  & \textbf{9.79} & 9.89 & \textbf{9.53} & 9.67 & 9.95 \\
\bottomrule
\end{tabular}}
\end{table*}

\section{Results}
\label{sec:results}

\subsection{Within-Dataset Regression}
\label{subsec:within}

Table~\ref{tab:within} reports within-dataset MAE for MLP-small
alongside the mean-predictor baseline (MAD, computed directly
from each label distribution).
All models nominally exceed the baseline on every task, but the
margins vary sharply. Coswara age is the only target with clear
signal (best/MAD\,=\,0.81, \HeAR{} $-0.46$\,yr over \MtwoD{}),
with CoughVID age weaker (best/MAD\,=\,0.93). By contrast, all
four CIDRZ targets sit at or near the chance floor (best/MAD
between 0.92 and 0.99; age within 1\% of the baseline), so the
FM embeddings carry no usable patient-level signal for this
clinical cohort under linear probing. The clinical-score targets
additionally show near-zero inter-model spread (TB: $\leq$0.004,
X-ray: $\leq$0.012\,MAE), consistent with a shared representational
ceiling rather than any model recovering true clinical variation.

\subsection{Generative vs.\ Contrastive Pretraining}
\label{subsec:gen}

\operaGT{} is favored over \operaCT{} on age regression in all
three datasets (CIDRZ $10.49$ vs.\ $10.52$\,yr, Coswara $10.16$
vs.\ $10.25$\,yr, CoughVID $9.62$ vs.\ $9.79$\,yr; 3/3 directions).
The CIDRZ gap ($0.03$\,yr) lies within one seed std, whereas the
Coswara and CoughVID gaps ($0.09$ and $0.17$\,yr) exceed seed
variance, so the trend is driven by the two larger datasets.
This is consistent with the breath-regression finding
of~\citet{operabenchmark}, suggesting the advantage extends
beyond breath audio.

\subsection{Regression Head Comparison}
\label{subsec:heads_res}

MLP-small wins in 23 of 30 model$\,\times\,$task combinations
(Table~\ref{tab:full_heads}), outperforming linear probing by up
to $0.38$\,yr (\HeAR{} on Coswara).
Full MLP overfits on CIDRZ ($N_{\text{train}}$\,=\,669):
\MtwoD{} degrades $+0.53$\,yr vs.\ MLP-small, driven by a
$\sim$22,000:1 parameter-to-sample ratio in the 3840-d hidden layer.
On larger CoughVID ($N_{\text{train}}$\,=\,3050), full MLP recovers
and \operaGT{} achieves its best result ($9.53$\,yr), revealing a
dataset-size $\times$ head-capacity trade-off as a deployment guideline.

\subsection{Cross-Dataset Generalisation}
\label{subsec:cross}

Table~\ref{tab:cross} reports cross-dataset age MAE for the best
model per transfer direction.
Transfer succeeds only when CIDRZ is the \emph{target}:
CoughVID$\,\to\,$CIDRZ achieves a negative gap ($-0.17$\,yr) and
Coswara$\,\to\,$CIDRZ is essentially flat ($+0.03$\,yr), both led
by \operaCE{}, showing that large-scale web-collected data can
substitute for scarce clinical training data.
The reverse fails: CIDRZ$\,\to\,$Coswara degrades $+2.43$\,yr
($+26.6\%$) and CIDRZ$\,\to\,$CoughVID $+0.94$\,yr, indicating
that small clinical populations do not generalise to large
crowdsourced ones.
\HeAR{} is best only in the three degrading directions
(gaps $\ge +0.94$\,yr), where it degrades least rather than
transferring well, so we do not read this as a pretraining-driven
advantage; the only non-degrading transfers are led by \operaCE{}.

\begin{table}[t]
\centering
\caption{Cross-dataset age generalisation (MLP-small, best model
per row). Gap\,=\,cross$\,-\,$within MAE of the same model.
\textbf{Bold} = negative gap (cross-dataset outperforms in-domain).}
\label{tab:cross}
\setlength{\tabcolsep}{4pt}
\small
\resizebox{\columnwidth}{!}{%
\begin{tabular}{llccc}
\toprule
Train $\to$ Test & Model & Cross & Within & Gap \\
\midrule
CoughVID $\to$ CIDRZ   & \operaCE{} & \textbf{10.34} & 10.51 & $-0.17$ \\
Coswara $\to$ CIDRZ    & \operaCE{} & 10.54          & 10.51 & $+0.03$ \\
Coswara $\to$ CoughVID & \operaCT{} & 10.42          &  9.79 & $+0.63$ \\
CoughVID $\to$ Coswara & \HeAR{}    & 10.05          &  9.12 & $+0.94$ \\
CIDRZ $\to$ CoughVID   & \HeAR{}    & 10.54          &  9.61 & $+0.94$ \\
CIDRZ $\to$ Coswara    & \HeAR{}    & 11.55          &  9.12 & $+2.43$ \\
\bottomrule
\end{tabular}}
\end{table}

\subsection{Low-Data Regime}
\label{subsec:lowdata}

\begin{figure}[t]
\centering
\includegraphics[width=\columnwidth]{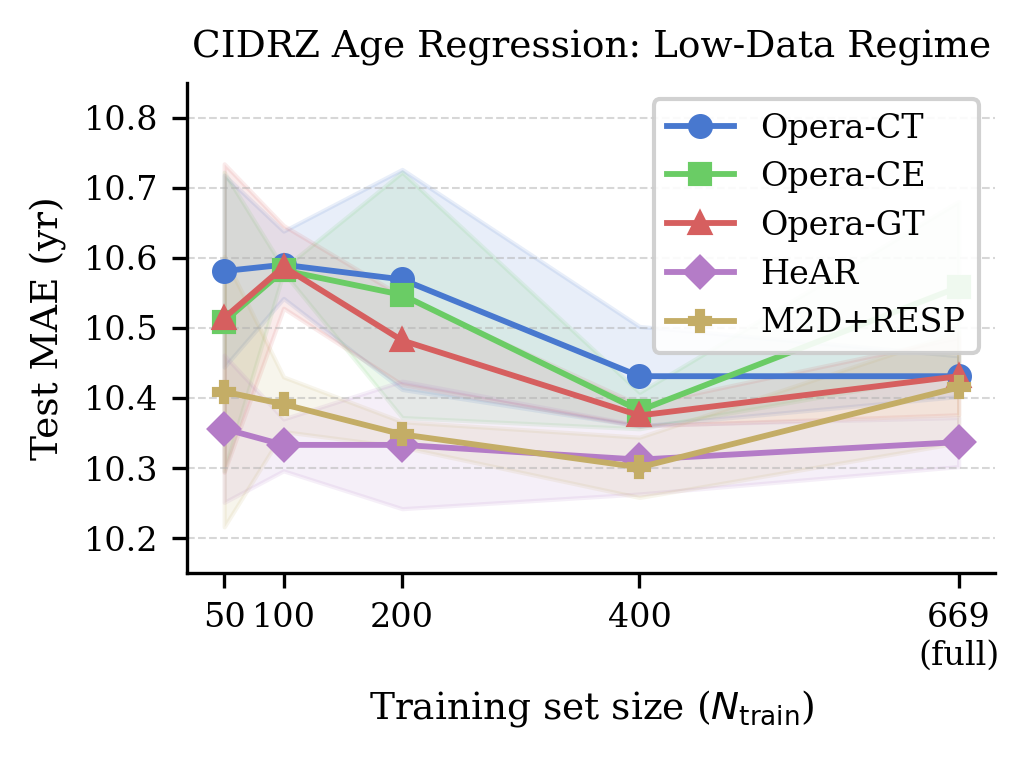}
\caption{CIDRZ age MAE vs.\ training set size (MLP-small,
3-seed mean\,$\pm$\,std). \HeAR{} and \MtwoD{} plateau by
$N$\,=\,100; OPERA models stabilise at $N$\,=\,400.}
\label{fig:low_data}
\end{figure}

Figure~\ref{fig:low_data} shows CIDRZ age MAE as a function of
training set size (MLP-small, 3 seeds).
\MtwoD{}, whose pretraining corpus does not include CIDRZ, shows
flat curves across all training sizes, indicating that large-scale
pretraining can produce embeddings that require minimal labeled
data for regression. \HeAR{} behaves identically, reaching within
$0.02$\,yr of its $N$\,=\,669 performance at $N$\,=\,50, consistent
with this account; on CIDRZ, however, we cannot exclude a
contribution from possible pretraining overlap.
OPERA models, also CIDRZ-clean, show substantially higher variance
at $N$\,=\,50 (std up to $\pm$0.22\,yr) and continue improving to
$N$\,=\,400, consistent with their smaller, respiratory-only
pretraining corpus.

\section{Conclusion}
\label{sec:conclusion}

We presented the multi-model, multi-target cough regression
benchmark evaluating five respiratory acoustic FMs across six targets,
three head architectures, and six cross-dataset transfer conditions.
Four findings carry practical implications.
First, MLP-small is the preferred head for frozen FM embeddings,
outperforming linear probing in 23 of 30 cases while avoiding
overfitting of full-width MLPs on small clinical datasets, though
signal strength varies substantially across tasks (best/MAD
ranging from 0.81 to 0.99), indicating that head choice matters
most when the task has learnable structure.
Second, generative pretraining (\operaGT{}) is favored over
contrastive pretraining (\operaCT{}) on age regression in all three
datasets (3/3 directions), though the effect is driven by Coswara
and CoughVID, as the CIDRZ margin lies within seed variance; this
extends a pattern observed for breath audio and is consistent with
the hypothesis that reconstruction-based objectives capture more
physiological variation, though the mechanism warrants further study.
Third, cross-dataset transfer is strongly asymmetric and large
web-collected data generalises to small clinical populations with
no performance loss while the reverse fails severely ($+2.43$\,yr,
81$\times$ larger gap), likely reflecting the narrow demographic
range of clinical cohorts relative to global crowdsourced corpora.
Fourth, \MtwoD{} reaches near-full performance at only $N$\,=\,50
labeled samples (with \HeAR{} behaving similarly) while OPERA models
require $N$\,=\,400, indicating that pretraining-corpus diversity
rather than architecture drives low-data regression; we anchor this
on the CIDRZ-clean \MtwoD{}, as \HeAR{}'s low-data CIDRZ result may
benefit from pretraining overlap.
\textbf{Limitations and future work.}
This study uses 2\,s clips. Cross-dataset transfer is reported for
age only because age is the sole target shared across two or more
datasets (BMI, X-ray abnormality, and TB probability appear only in
CIDRZ), so transfer is undefined for these targets rather than
omitted. All conclusions are likewise scoped to the
frozen-embedding, shallow-probe regime: the benchmark characterises
what FM embeddings encode, not the performance ceiling achievable
with fine-tuning, which was not evaluated and may further reduce the
transfer gap on small clinical datasets. Future work should extend
to additional targets, fine-tuning regimes, and attention-pooling
heads.

\section*{Impact Statement}

This paper presents work whose goal is to advance the field of
Machine Learning. There are many potential societal consequences
of our work, none of which we feel must be specifically highlighted
here.

\bibliography{example_paper}
\bibliographystyle{icml2026}

\end{document}